\documentclass[10pt,twocolumn,letterpaper]{article}

\usepackage[pagenumbers]{cvpr} 

\usepackage{wrapfig}
\usepackage{multirow}
\usepackage{pifont}
\usepackage{booktabs}
\usepackage[table]{xcolor}

\newcommand{\acronym}{Human$_0$}
\newcommand{\datasetname}{PH$^{S}$D}

\definecolor{cvprblue}{rgb}{0.21,0.49,0.74}
\usepackage[pagebackref,breaklinks,colorlinks,allcolors=cvprblue]{hyperref}

\def\paperID{2160} 
\def\confName{CVPR}
\def\confYear{2026}

\title{In-N-On: Scaling Egocentric Manipulation with \textbf{in}-the-wild and \textbf{on-task} Data}

\author{
Xiongyi Cai$^{*}$ \quad Ri-Zhao Qiu$^{*\dagger}$ \quad Geng Chen \quad Lai Wei \\
Isabella Liu \quad Tianshu Huang \quad Xuxin Cheng \quad Xiaolong Wang \\
UC San Diego\\
$^{*}$ Equal Contribution \quad $^{\dagger}$ Project Lead\\
\url{https://xiongyicai.github.io/In-N-On}
}

\begin{document}

\twocolumn[{%
\renewcommand\twocolumn[1][]{#1}%
\maketitle

\begin{center}
    \centering
    \captionsetup{type=figure}
    \includegraphics[width=\linewidth]{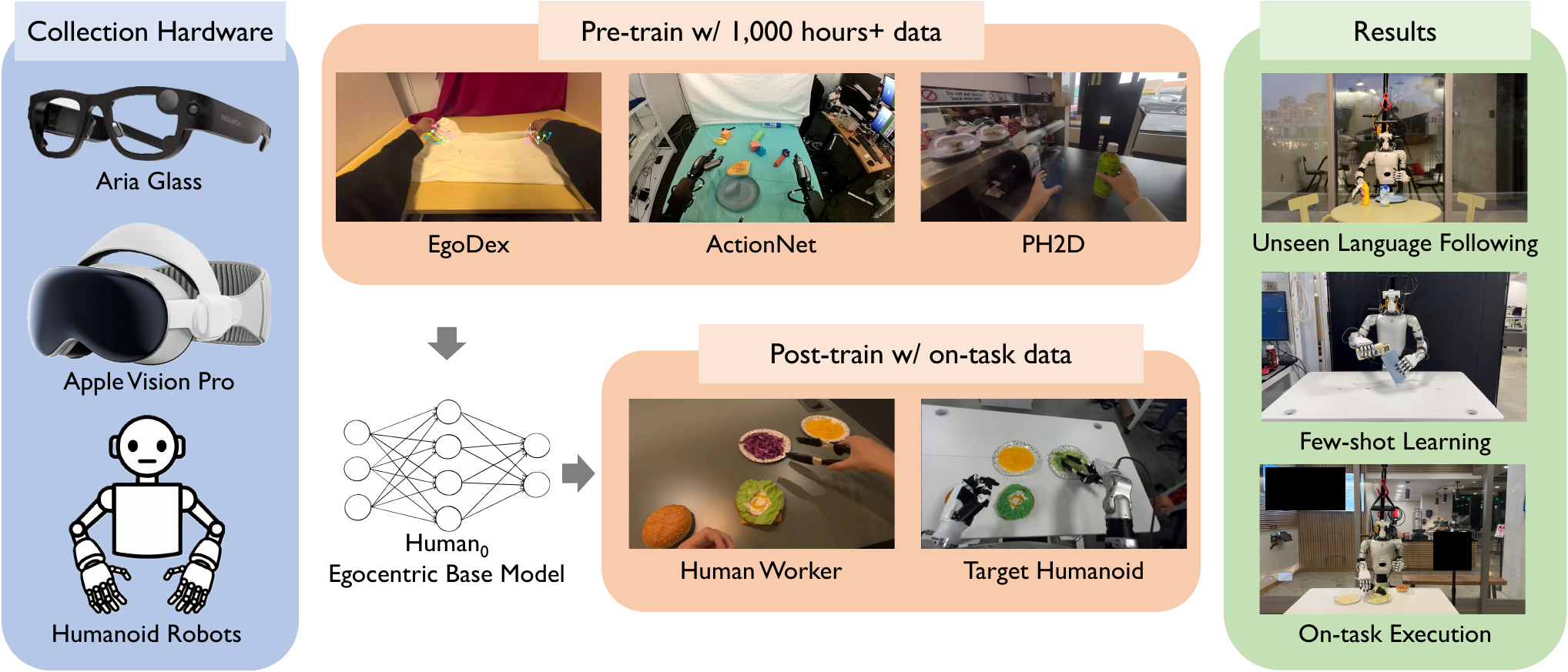}
    \caption{This paper investigates large-scale pre-training and post-training with egocentric human data. We curate a large-scale \textbf{P}hysical \textbf{H}uman-humanoid\textbf{S} \textbf{D}ataset, dubbed \datasetname{}, to train a base model to model egocentric human-humanoid behavior. Empirically, we show that \acronym{} achieves several interesting properties, including strong language following of instructions unseen in robot data, few-shot execution, and improved on-task performance. }
    \label{fig:teaser}
\end{center}
}]

\begin{abstract}
Egocentric videos are a valuable and scalable data source to learn manipulation policies. However, due to significant data heterogeneity, most existing approaches utilize human data for simple pre-training, which does not unlock its full potential. This paper first provides a scalable recipe for collecting and using egocentric data by categorizing human data into two categories: \textbf{in-the-wild} and \textbf{on-task} alongside with systematic analysis on how to use the data. We first curate a dataset, \datasetname{}, which contains over 1,000 hours of diverse in-the-wild egocentric data and over 20 hours of on-task data directly aligned to the target manipulation tasks. This enables learning a large egocentric language-conditioned flow matching policy, \acronym{}. With domain adaptation techniques, \acronym{} minimizes the gap between humans and humanoids. Empirically, we show \acronym{} achieves several novel properties from scaling human data, including language following of instructions from only human data, few-shot learning, and improved robustness using on-task data.

\end{abstract}    
\section{Introduction}

The robot manipulation community has recently witnessed great progress in learning from real robot demonstrations~\citep{black2024pi_0,intelligence2025pi_,liu2024rdt,barreiros2025-lbm,kim2025-OpenVLAOFT,zhao2024-aloha-unleashed}. Behind the curtain are novel algorithms~\citep{intelligence2025pi_} and large-scale robot data~\citep{o2024-open-x,black2024pi_0}, which enable dexterous and long-horizon tasks~\citep{barreiros2025-lbm}. However, existing foundational manipulation policies still lack \textit{robust real-world generalizability} compared to their counterparts in LLM~\citep{achiam2023-gpt} or self-driving~\citep{TeslaFSD} that are trained on much larger-scale data.

In search of a novel data source to fuel model training, researchers have turned to cross-embodiment learning from different robots~\citep{o2024-open-x,kim2024openvla,black2024pi_0}, and, more recently, to human data~\citep{ma2024-nymeria,grauman2022-ego4d,grauman2024-ego-exo4d,banerjee2025-hot3d}. Intuitively, humans are naturally the most prominent physical embodiment compared to other morphologies that can easily manipulate daily objects. Thus, learning from human data has been studied for over a decade. Modular methods learn affordance~\citep{wang2017-binge,mendonca2023-human-world,bahl2023affordances} and plan robot manipulation in a model-based fashion~\citep{li2024-okami,xiong2021learning,qin2022-dexmv}. More recently, advances in computer vision have enabled precise finger keypoint tracking to generate human data with action labels. Recent methods~\citep{kareer2025egomimic,qiu2025humanoid,luo2025being,yang2025egovla,niu2025human2-locoman,lepert2025masquerade,qin2022-dexmv,li2025-scalable} have shown that such high-quality data can be directly used for end-to-end training, which has the potential to be easily scaled up.

However, the vast amount of human data also leads to significant data heterogeneity. Existing human datasets are very diverse - ranging from daily activities such as walking and dancing~\citep{grauman2022-ego4d}, long-horizon kitchen activities~\citep{Damen2018-epickitchen}, and even sitcoms~\citep{wang2017-binge}. To address such heterogeneity (or misalignment between embodiments), some methods propose new algorithms to use intermediate representations such as object pose~\citep{li2024-okami} or affordance~\citep{bahl2023affordances} to learn from these \textbf{in-the-wild} datasets. On the other hand, recent end-to-end approaches~\citep{yang2025egovla,luo2025being,bi2025h} have resolved to scaling up pre-training with human data, and then fine-tuning with robot data. This approach is usually sub-optimal due to catastrophic forgetting~\cite{french1999-catastrophic,hancock2025-vlavlm} from simple fine-tuning with highly heterogeneous data.

On the other hand, recent methods~\citep{qiu2025humanoid,kareer2025egomimic,wang2024-dexcap,xu2025-dexumi,objectsdexwild} have also focused on collecting \textbf{on-task} human data. Instead of recording casual activities, on-task data collection focuses on curating human demonstrations on  the same tasks that robots will be working on ({\it e.g.,} recorded by actual human workers). Compared to in-the-wild data, on-task data are more task-oriented, in-domain, and segmented well. These factors ensure good alignment to the target deployment distribution, which has been empirically shown to enable direct co-training of mixed humans and robot data~\citep{qiu2025humanoid,kareer2025egomimic,objectsdexwild,xu2025-dexumi} to mitigate catastrophic forgetting from pre-training.

The goal of this paper is to show that it is important to use both in-the-wild and on-task data to unlock the full potential of human data: in-the-wild data is easy to collect and diverse, but it may be only suitable for bootstrapping a base model. In contrast, on-task data is more well-aligned with the target distribution but often smaller in magnitude.

To this end, we investigate the boundary between these two paradigms. Our insight is to use \textbf{in}-the-wild data \textbf{and} \textbf{on}-task data for pre-training and post-training. With language annotations and a unified human-centric action space~\citep{qiu2025humanoid}, this enables learning of a large egocentric language-conditioned flow matching policy, \acronym{}. In addition to scaling up egocentric training data, we perform systematic study to reveal that na\"ive data mixing leads to hidden states that discriminate robot and human inputs. \acronym{} adopts domain adaptation technique to improve hidden states to fully utilize human training data.

We evaluate \acronym{} on a real Unitree H1 humanoid and a Unitree G1 humanoid equipped with 5-fingered dexterous hands. Empirically, the pre-training and post-training for \acronym{} achieve several novel properties, including language following of instructions that are unseen in the robot training data and few-shot learning, which is validated by systematic ablation studies. In particular, we studied a task, \textit{fast food worker}, where data can be collected at a low marginal cost from real food-industry worker. We show how the on-task data collected for this practical scenario improves policy robustness drastically.

In sum, our contributions are,
\begin{itemize}
    \item A large-scale human-humanoid dataset, \datasetname{}, that provides data recipe for pre-training and post-training an egocentric model. We plan to open-source the dataset.
    \item A base egocentric manipulation model, \acronym{}, which is augmented with the domain adaptation technique and applicable to many egocentric bimanual embodiments. The weights will be open-sourced.
    \item Extensive experimental results with demonstrations of language following and few-shot learning on real humanoid robots.
\end{itemize}

\section{Related Work}

\textbf{Large-scale Manipulation Models.}
Recent advances in vision-language-action (VLA) models have shown promising progress in robotic manipulation tasks, with a growing emphasis on models' robustness and generalization.
Building upon early efforts in learning from real-robot demonstrations~\citep{chi2023-DP,zhao2023-ACT}, recent methods~\citep{kim2024openvla, black2024pi_0, liu2024rdt, intelligence2025pi_, barreiros2025-lbm, rdt2, zheng2025-xvla} explored how to scale up robot manipulation policy training with more data. The advances happened both in the modeling regime and the data regime. In the context of modeling, VLAs extend vision-language models (VLMs) or large-language models (LLMs) with action decoders to make use of pre-trained knowledge infused in VLMs. More recently, \citet{intelligence2025pi_} also proposed a new paradigm to make the training process more data-efficient. On the other hand, data is important for scaling up the manipulation model. Notably, many large manipulation models~\citep{liu2024rdt,black2024pi_0} rely on cross-embodiment learning~\citep{o2024-open-x}, where a model designs its architecture specifically to work with data from multiple robot embodiments. However, even with cross-embodiment learning, the magnitude of available data is still significantly smaller compared to counterparts in language or vision models. Current manipulation models are data-hungry for more generalizability.

\begin{figure*}[t]
    \centering
    \includegraphics[width=1.\textwidth]{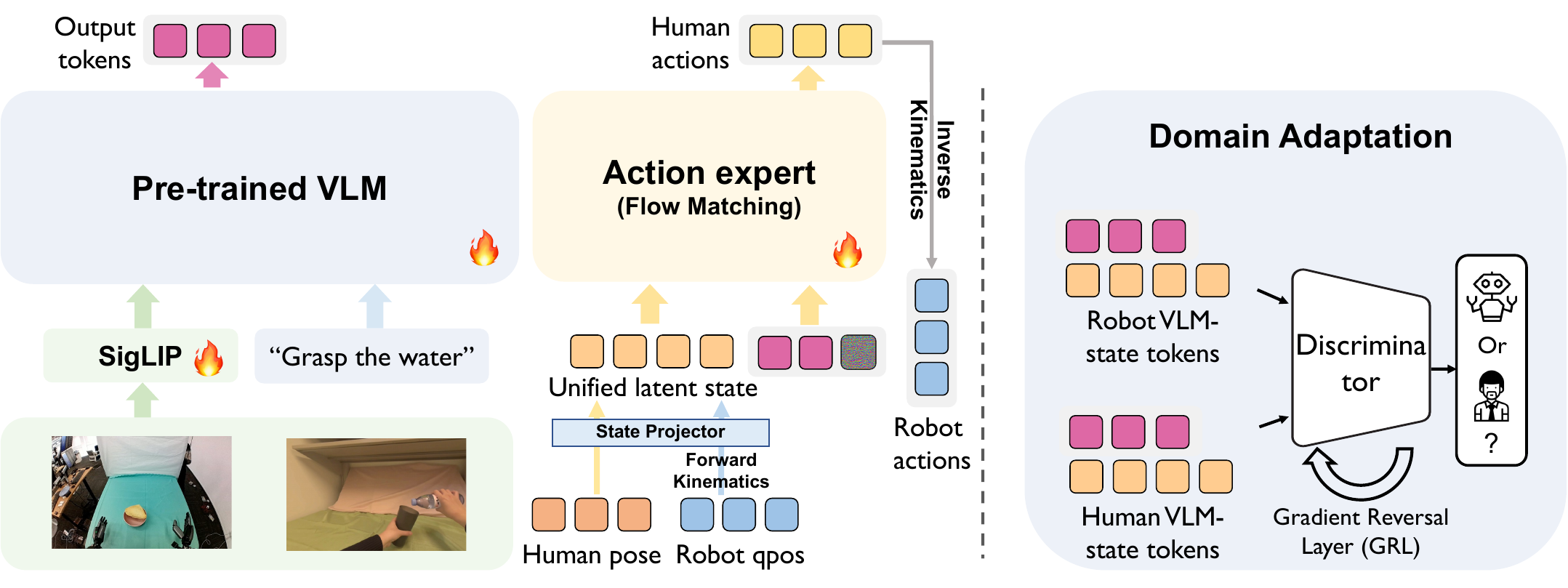}
    \caption{
    Method overview. Our approach follows a two-stage training recipe: (1) pre-training on large-scale in-the-wild human and robot data that are mapped into a unified human-centric state-action space; and (2) on-task post-training using task-aligned human and robot demonstrations. To bridge the embodiment gap, We employ a domain-adversarial discriminator that takes SigLIP visual features and action-state embeddings as input and predicts whether a sample is from human or robot data. Through gradient reversal, this encourages the policy's encoders to produce embodiment-invariant representations, enabling effective transfer between human and robot observations.
    }
    \vspace{-10pt}
    \label{fig:overview}
\end{figure*}

\textbf{Learning from Human Videos.}
Learning robot policies from human videos has been an active research direction, driven by the availability of large-scale human data.
Early efforts~\citep{nair2022r3m, radosavovic2023real, ma2022vip} focused on leveraging human videos to pre-train visual representations that are better suited for downstream manipulation policy learning; or to leverage human videos to learn intermediate representations such as affordance~\citep{bahl2023affordances}.
Beyond pre-training on visual tasks for improved initializations, other works~\citep{wang2023mimicplay, bharadhwaj2024towards, xiong2021learning, wen2023any, bharadhwaj2024track2act, bahl2023affordances} attempt to use human data directly for downstream tasks such as point tracking~\citep{bharadhwaj2024track2act,wen2023any,li2025-novaflow}, and high-level planner~\citep{wang2023mimicplay}, which are then used to guide robot action prediction. 

\textbf{End-to-end Learning Manipulation Policies from Human.}
An increasing number of works have started to investigate scaling manipulation in an end-to-end manner by leveraging human demonstrations~\citep{kareer2025egomimic, bi2025h, qiu2025humanoid, yuanmotiontrans, zhu2025emma, objectsdexwild, punamiyaegobridge,li2025-scalable}. They either use diverse \textbf{in-the-wild} data for pre-training~\citep{bi2025h,yang2025egovla,luo2025being,li2025-scalable} or \textbf{on-task} data for co-training. Notably, \citet{bi2025h,luo2025being,li2025-scalable} have shown pre-training with human data leads to improved generalizability; \citet{lepert2025masquerade} apply modular vision modules to edit human videos to match robot videos to reduce visual gaps. Concurrently, EMMA~\cite{zhu2025emma} learns a mobile manipulation policy using human data. However, there has yet to be an attempt to explore both in-the-wild  data and on-task data to cover both pre-training and post-training stages. This paper aims to bridge such a gap by prescribing a recipe for data curation, an end-to-end large egocentric manipulation base model, and algorithmic advances to improve the model.

\section{Method}

This paper discusses models and data recipes for pre-training and post-training a base model for egocentric manipulation, as well as analysis of design decisions made to create the recipes. Sec.~\ref{sec:dataset} describes the curation process for a large human-humanoid dataset. Sec.~\ref{sec:model} discusses the design choices for the base model, including data mixture and domain adaptation.

\subsection{\datasetname{}: Physical Humans-Humanoids Dataset}
\label{sec:dataset}

Many human datasets~\cite{qiu2025humanoid,hoque2025egodex,fan2023-arctic,zhan2024-oakink2,ma2024-nymeria,wang2023-holoassist} and egocentric robot datasets~\cite{zhao2025-humanoid-everyday,fourier2025-actionnet} exist. Naturally, the formats of the human dataset are similar - most existing datasets focus on tracking head, wrists, and fingers poses. However, humanoid hardwares, or robot hardware in general, are far from convergence. Therefore, these publicly available egocentric robot datasets are vastly different - kinematics, DoFs, and mechanical configurations can differ. The difference in state-action space in each dataset hinders scaling up the training size.

To tackle this, existing methods attempted to design physically explainable state-action space~\cite{liu2024rdt} or operate in the unified latent space~\cite{wang2024-hpt}. However, scaling data in the same state-action space remains the most explainable and effective way~\cite{rdt2,qiu2025humanoid}. For egocentric manipulation, this paper advocates the human representation for learning, as humans are the most prevalent embodiment and are the sources of biological inspiration for bimanual robot designs.

To this end, this paper defines a unified human-centric state-action space. We then implement a software suite of robot IK/FK (Inverse Kinematics and Forward Kinematics), and hand retargeting algorithms to differentiably convert human and humanoid data from/to our unified space. Finally, we curate and process data from multiple sources into a unified format for training.

\begin{figure}[t]
  \centering
  \includegraphics[width=\linewidth]{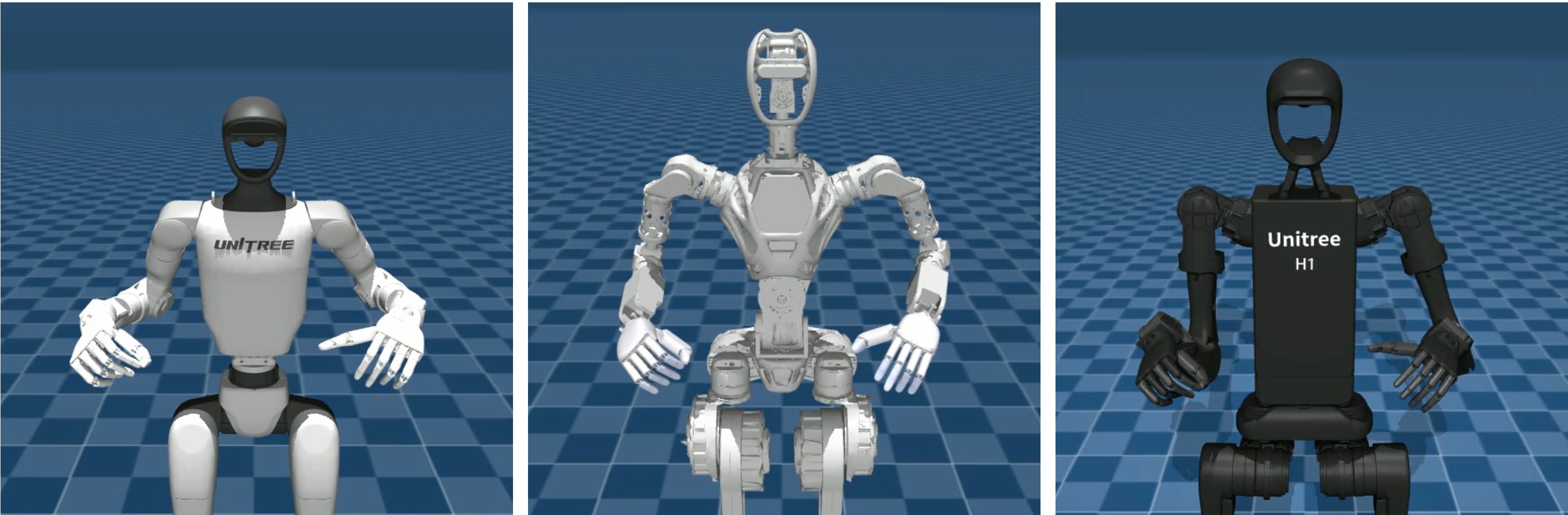}
  \caption{Our retargeting software suite supports retargeting different humanoids from/to the human-centric representation. Figure demonstrates retargeting from the same human action to different humanoids in MuJoCo~\cite{todorov2012-mujoco}. The code will be released.}
  \label{fig:retargeting}
\end{figure}

\textbf{Unified human-centric state-action space.} Following human activities datasets~\cite{ma2024-nymeria,hoque2025egodex,wang2023-holoassist} and Human-Humanoid co-learning~\cite{qiu2025humanoid}. We design the state-action space to have the following elements.
\begin{itemize}
    \item $\mathbf{T_{head}} \in \mathbf{SE}(3)$. We parameterize head poses as the base transformation with rotation and translation. Compared to previous work~\cite{qiu2025humanoid}, this further encodes translation to support potential applications such as whole-body loco-manipulation. Current egocentric human data is usually collected by wearable devices mounted on operators' head. Hence the localization frame is usually modeled as world-head transformation.
    \item $\mathbf{T_{Lwrist}}, \mathbf{T_{Rwrist}} \in \mathbf{SE}(3)$. The wrist poses are modeled as relative to the head pose. Though there is inevitable physical difference ({\it e.g.,} height) among different data collectors, such a difference can be neglected as the training data scales.
    \item $\mathbf{P_{Lfinger}}, \mathbf{P_{Rfinger}} \in \mathbb{R}^{3 \times 5}$: we model finger motions as fingertip keypoints, as all well-established optimization-based finger retargeting algorithms~\cite{handa2020-dexpilot,qin2023-anyteleop,cheng2024-opentv} directly use fingertip keypoints with scaling factors.
    \item $\mathbf{D_{Lgripper}}, \mathbf{D_{Rgripper}} \in \mathbb{R}$: optionally, for bimanual robots equipped with parallel grippers, we map gripper distance to human thumb-index fingertip distance. Note that $\mathbf{P_{Finger}}$ is sufficient for computing $\mathbf{D_{gripper}}$.
\end{itemize}

\textbf{Retargeting Software Suite.} To make it easy for us and the research community to explore the human-centric state-action space for egocentric manipulation, we implement an IK/FK and retargeting software suite based on Pinocchio~\cite{carpentier2019-pinocchio}. Our suite converts the robot joint positions from/to human-centric space (Fig.~\ref{fig:retargeting} visualizes same human action retargeted to different humanoids in the accompanying MuJoCo~\cite{todorov2012-mujoco} simulator in our suite). Therefore, as long as a released humanoid manipulation dataset provides joint readings, it can be used for training. We hope to facitate large-scale egocentric learning on humanoid robot. The code and assets will be open-sourced.

\textbf{Aggregating In-the-wild datasets for Pre-training.} Building on our retargeting framework, we use EgoDex~\citep{hoque2025egodex}, Fourier ActionNet~\citep{fourier2025-actionnet}, and PH2D~\citep{qiu2025humanoid} for pre-training. Note that with our software suite, our method also applies to future and concurrent dataset~\cite{zhao2025-humanoid-everyday}.
\begin{itemize}
    \item EgoDex~\citep{hoque2025egodex} contains 800+ hours of skill-rich human demonstrations, which were collected using multiple Apple Vision Pros. It contains 6dof head pose, wrist pose, and finger keypoints.
    \item The ActionNet dataset~\citep{fourier2025-actionnet} contains over 100 hours of humanoid demonstrations - most of which were done on the Fourier GR1T1 robot embodiment equipped with bimanual Fourier 5-fingered 6-DoF dexterous hands.
    \item The PH2D~\citep{qiu2025humanoid} dataset contains human and humanoid demonstrations of various tasks. Similar to EgoDex~\citep{hoque2025egodex}, PH2D also collected human data with Apple Vision Pro, which can be processed in a similar manner. The humanoid data (collected on Unitree H1 with 5-fingered Inspire hands) are also processed by our software suite.
\end{itemize}

\textbf{Data for Post-training.} To ensure high-quality hand poses in our on-task datasets, we use commercial-grade data collection devices, including Apple Vision Pro and the Meta Aria Glass. Both Vision Pro and Meta Aria glass provide head poses, wrist poses and dense fingertip keypoint predictions. The human dataset also includes 2D keypoint projection, as we hope our released data can also help other approaches such as generative inpainting~\cite{lepert2025masquerade,zhang2023-controlnet}. The robot dataset are collected using Apple Vision Pro with OpenTV~\cite{cheng2024-opentv}. Visualizations and projections of these datasets can be found in the supplementary material.

\subsection{\acronym{}: Foundational Egocentric Base Model}
\label{sec:model}

\subsubsection{Architecture}

While the pre-training and post-training recipes proposed in this paper are model-agnostic, we adopt a language-conditioned flow matching model~\citep{black2024pi_0}. Specifically, a SigLIP-based vision module extracts visual tokens $v \in \mathbb{R}^{L \times C}$, where $L$ is the number of patches and $C$ is the embedding dimension. The SigLIP encoder provides strong alignment between visual inputs and text, enabling downstream instruction grounding. Visual tokens are then combined with text embeddings $n \in \mathbb{R}^{T \times C}$ to form a joint multi-modal representation, which is further processed in the transformer blocks to propagate cross-modal context.

To use the human-centric representation, we use lightweight MLPs to encode input states and the output actions. For the input states, the physically interpretable human-centric state is projected to a pose latent $x \in \mathbb{R}^{C}$. We denote the latent tokens produced by the backbone transformer as
\begin{equation}
    z = \text{Transformers}(v, n, x), \quad z \in \mathbb{R}^C,
\end{equation}
which integrate information across modalities. Unless otherwise noted, we use the pre-trained checkpoint released by \citet{black2024pi_0} to initialize the model pre-training. Note that since the human-centric representations introduce larger vector sizes and different interpretations of each element at different indices, we swap out the original projection modules with different dimensions and random initialization.

\begin{figure}[t]
  \centering
  \begin{subfigure}{\linewidth}
    \centering
    \includegraphics[width=0.8\linewidth]{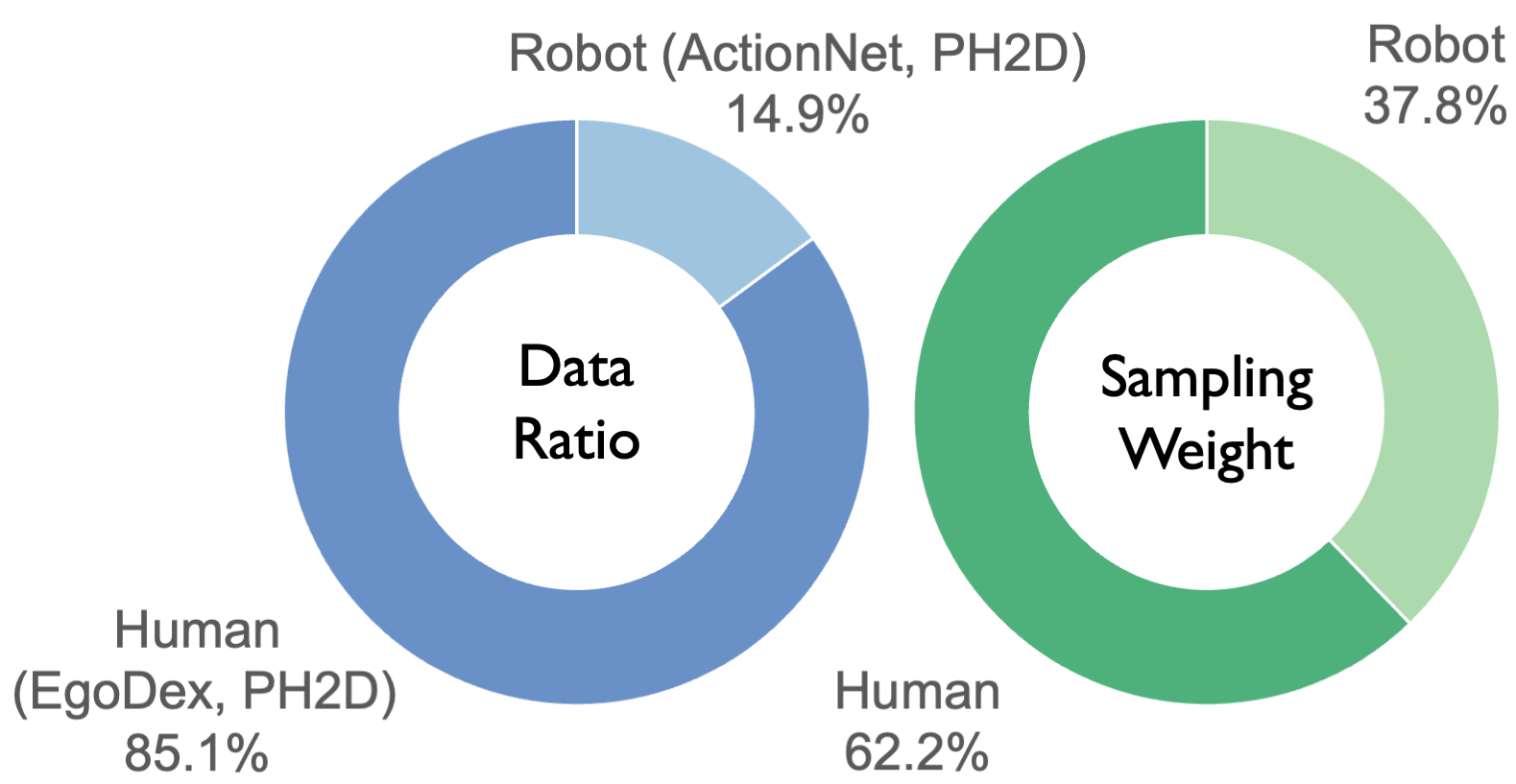}
    \caption{Data size ratio and sampling factor for \textbf{pre-training} data.}
    \label{fig:pretrain_weighing}
  \end{subfigure}

  \vspace{4pt}

  \begin{subfigure}{\linewidth}
    \centering
    \includegraphics[width=0.8\linewidth]{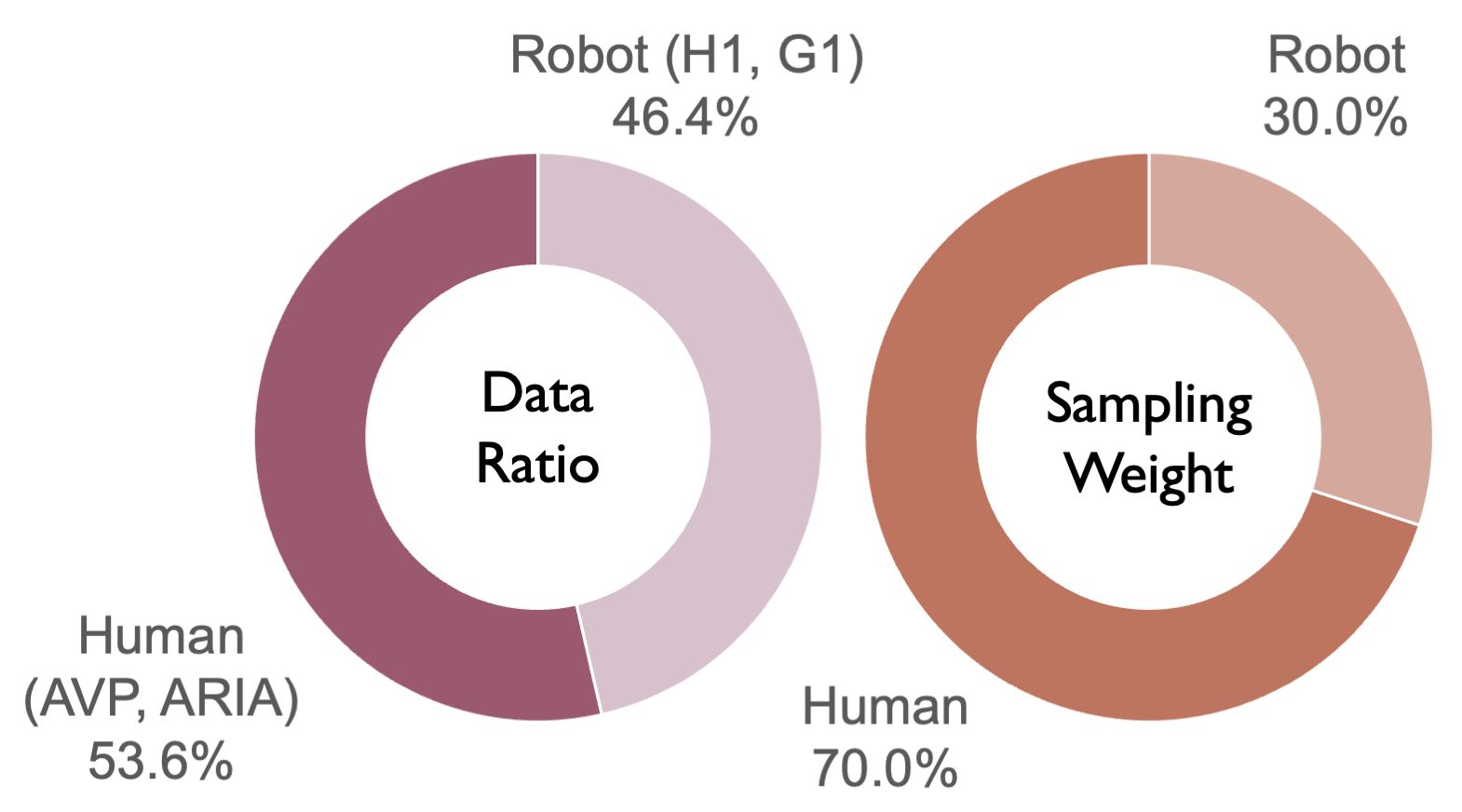}
    \caption{Data size ratio and sampling factor for \textbf{post-training} data.}
    \label{fig:posttrain_weighing}
  \end{subfigure}

  \caption{Data distributions and sampling factors for pre-training and post-training.}
  \vspace{-10pt}
  \label{fig:train_weighing}
\end{figure}

\subsubsection{Pre-training on Human and Robot Data}

We first pretrain the base model using over 1,000+ hours of mixed data from EgoDex~\citep{hoque2025egodex}, ActionNet~\citep{fourier2025-actionnet}, and PH2D~\citep{qiu2025humanoid}, covering rich egocentric human and robot manipulation scenarios. During this stage, the objective is to learn a unified vision–language–action prior that models human-like behaviors across different embodiments.

More concretely, let $a \in \mathbb{R}^C$ be the target action, the model is trained with a flow-matching objective in an end-to-end manner:
\begin{equation}
    \mathcal{L}_{\text{FM}} = 
\mathbb{E} \Big[ \, \| f_\theta^{\text{flow}}(z, a_t, t) - (a - u) \|_2^2 \, \Big]\,,
\label{eq:flow_matching_loss}
\end{equation}%
where the Gaussian noise vector $u \sim \mathcal{N}(0, I_C)$, time step $t \sim U(0, 1)$, and interpolated action $a_t = (1 - t)u + ta$. The $f_\theta^{\text{flow}}(z, a_t, t)$ represents the predicted flow vector, which points from the noisy sample towards the target. This pre-training stage equips the base model with broad visuomotor priors from the vast amount of human videos. In addition, it aligns the VLM originally trained on image-text data to model human behavior. The shared embodiment space provides a strong regularization for post-training, enabling effective transfer between human and robot manipulation.

\textbf{Data mixing recipe.} The distribution of the pre-training data is presented in Fig.~\ref{fig:pretrain_weighing}. Note that due to the overwhelming amount of human data in pre-training, we manually adjust the training data sampler ratio to balance and stabilize the training process.

\subsubsection{Post-training on Human and Robot Data}

During post-training, we focus exclusively on human and robot data collected for the task of interest. The goal is to refine the policy's language grounding and visuomotor control to match the distribution of real-world tasks, where data can be collected by actual human workers performing these real-world tasks. Thanks to our unified action space design, the training procedure follows Eq.~\eqref{eq:flow_matching_loss} precisely.

\textbf{Data mixing recipe.} The distribution of post-training data is presented in Fig.~\ref{fig:posttrain_weighing}. Compared to pre-training, our post-training dataset has considerably more robot data. Empirically, we found that sampling slightly more often ({\it e.g.,} $70\%$) from the human data helps preserve semantics in human data better. This finding is somewhat consistent with \citet{objectsdexwild}, which used an 8:2 sampling ratio to sample human data more often.

In summary, our pre-training and post-training process enables various interesting properties, including (1) language following of instructions unseen in robot data; (2) few-shot robot data learning with as few as 1 demonstration; and (3) improved robustness across related tasks.

\subsubsection{Domain Adaptation: Blurring the Line between Embodiments}

Ideally, our model should be embodiment-agnostic and process all egocentric data from a human-centric perspective. However, though we use image augmentation and human-centric representation with forward kinematics to provide regularization, the model can still learn to distinguish different embodiments, resulting in overfitting to a specific configuration.

\setlength{\intextsep}{8pt}
\setlength{\columnsep}{5pt}
\begin{wrapfigure}{r}{0.20\textwidth}
  \centering  
  \includegraphics[width=\linewidth]{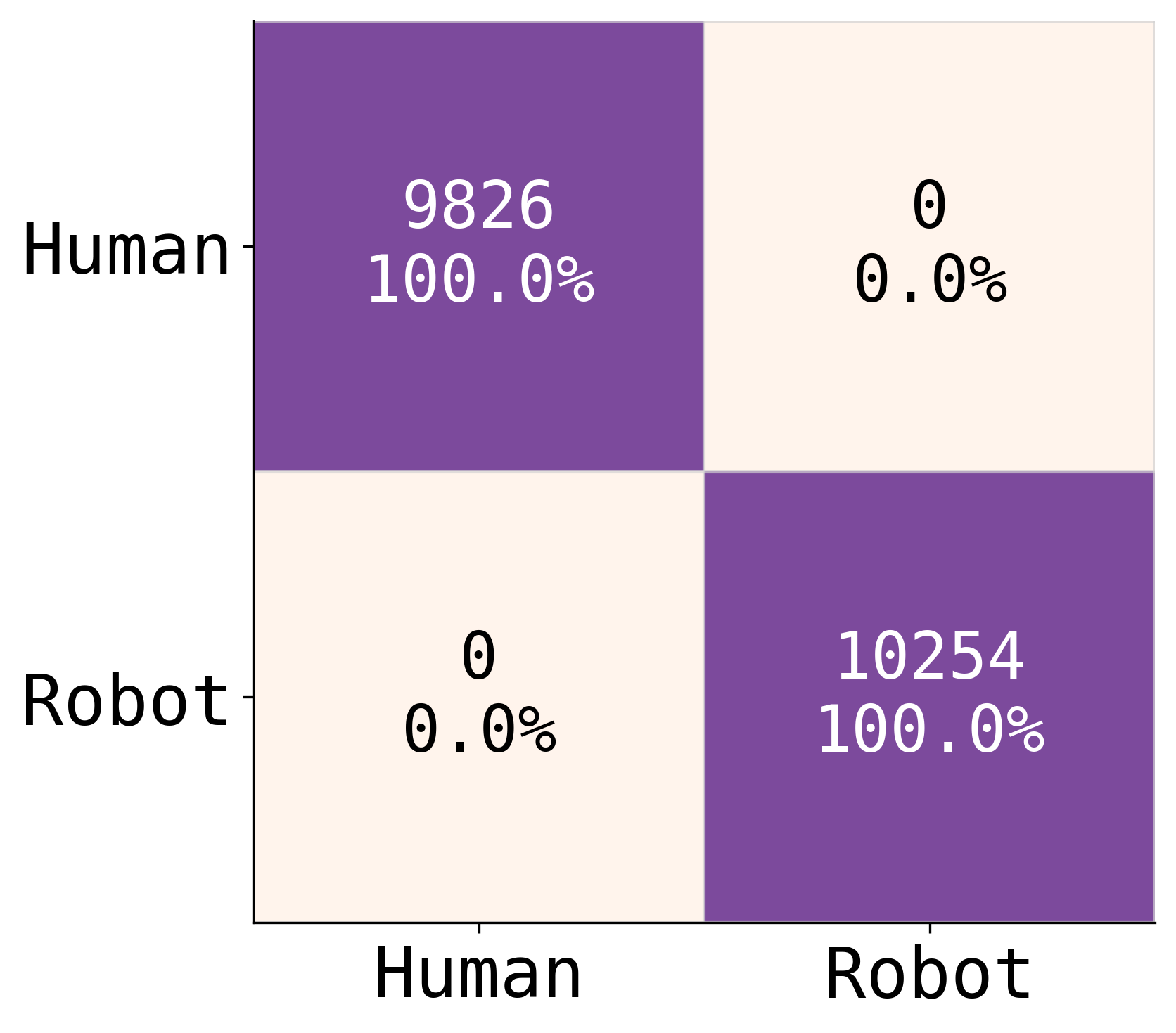}
  \vspace{-23pt}
  \caption{Confusion matrix obtained by linear probing intermediate features from vanilla model.}
  \label{fig:confusion_matrix_no_dis}
\end{wrapfigure}
To verify this, we first pre-train and post-train the model using vanilla denoising objective Eq.~\eqref{eq:flow_matching_loss}. Then, we perform a simple linear probing study. We train a simple MLP taking intermediate visual tokens and proprioceptive tokens as inputs. The training objective for the MLP is a binary classification problem, where it tries to predict if a set of concatenated visual and proprioceptive tokens belongs to human data or robot data. The results are shown in Fig.~\ref{fig:confusion_matrix_no_dis}. Surprisingly, on a held-out validation set, the simple MLP achieves 100\% success rate - suggesting that the model `cheats' by implicitly biasing features to recognize if the input is human or robot. (More technical details are given in the supplementary material).
\begin{figure*}[t]
    \centering
    \includegraphics[width=\textwidth]{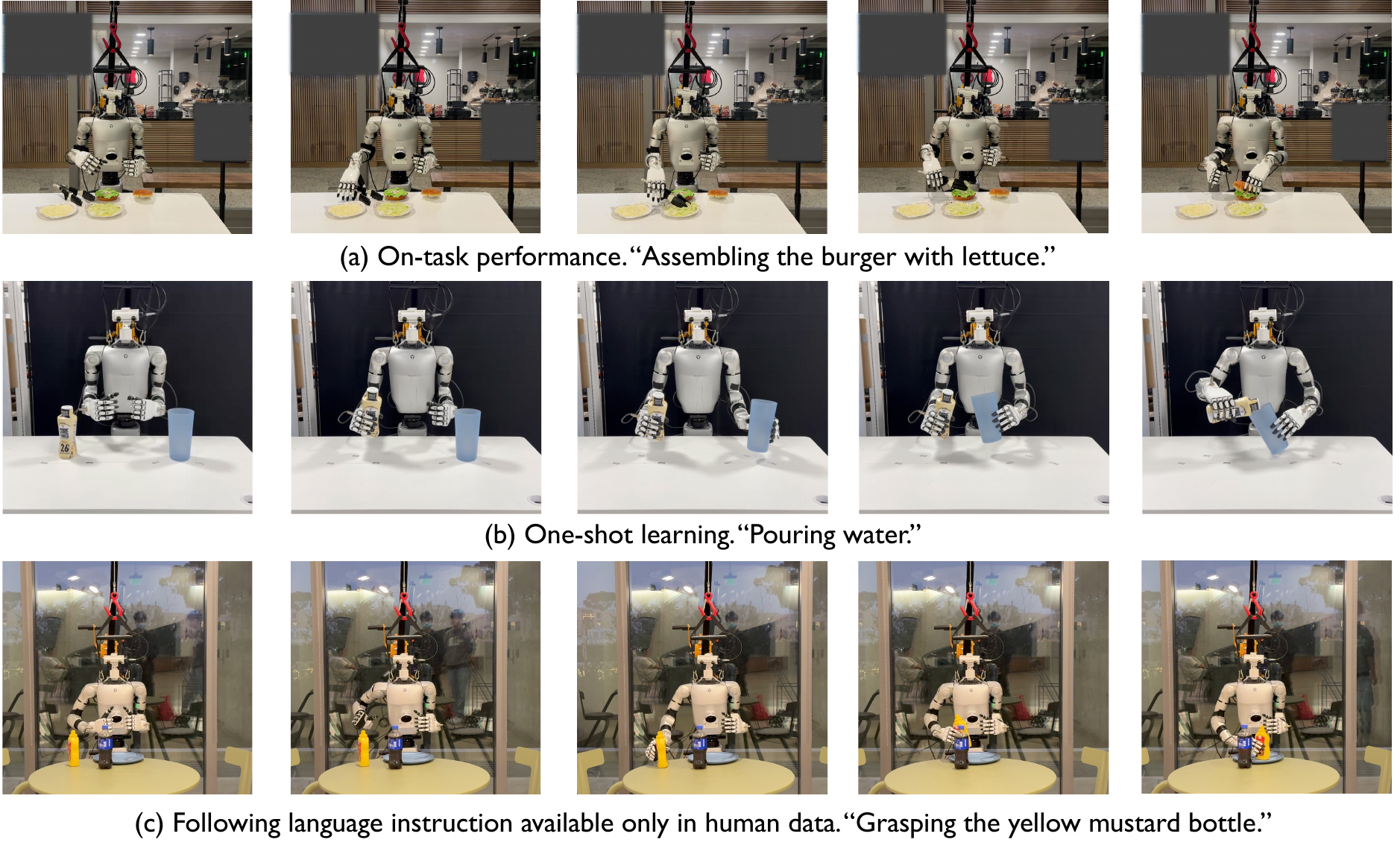}
    \vspace{-25pt}
    \caption{
        We task the robot to perform several manipulation tasks to evaluate few-shot learning, language instruction following, and robustness using on-task human data. Videos in the supplementary.
        (Top to bottom: burger assembly, pouring, and multi-object grasping).
    }
    \vspace{-10pt}
    \label{fig:real_robot}
\end{figure*}
To discourage the model from overfitting to specific visual cues or proprioceptive cues, we introduce a discriminator network~\citep{ganin2016-GRL}. Specifically, the network is tasked to classify the type of embodiment. Following \citet{ganin2016-GRL}, the network is modeled as a MLP that takes in intermediate features with Gradient Reversal Layer (GRL)~\citep{ganin2016-GRL} to discourage successful classification.

More specifically, the GRL is trained to differentiate between the feature encoding of human data and those of robot data. We concatenate the visual tokens $v$ from SigLIP encoder with the projected pose latent $x$ along the token dimension, and pass them though a attention head to obtain a feature vector:
\begin{equation}
    m = \text{Attn} \big ( \text{Concatenate}( v, x ) \big ), \quad m \in \mathbb{R}^C\,.
\end{equation}%
The feature vector $m$ is then passed through the discriminator MLP $D_\theta$ that predicts the input’s embodiment type. The discriminator is trained with binary cross-entropy loss:

\begin{equation}
    \mathcal{L}_{D}(\phi \mid \theta) = 
- \mathbb{E} \big[ \log D_\phi(m_h) \big]
- \mathbb{E} \big[ \log (1 - D_\phi(m_r)) \big],
\end{equation}

where $m_h$ and $m_r$ denote feature vectors obtained from human and robot data,respectively. With a GRL inserted between feature vectors and discriminator $D_{\phi}$, the optimization is adversarial: $D_{\phi}$ minimizes $\mathcal{L}_D$, while the backbone policy encoders $f_\theta$ maximizes it:

\begin{equation}
    \max_{\theta} \ \min_{\phi} \ \mathcal{L}_{D}(\phi, \theta)
\end{equation}

In other words, the GRL encouraging the upstream policy encoder to produce features that are invariant to the human-robot domain distinction. This adversarial setup promotes feature alignment across data domains and embodiments, enabling more effective transfer of manipulation behaviors between human demonstrations and robot.

\textbf{Final Loss.} Combining both flow matching L2 loss and the domain adaptation loss, the final training loss is given by
\begin{equation}
    \mathcal{L}_{\text{final}} = \mathcal{L}_{\text{FM}} + \lambda \cdot \mathcal{L}_{D}(\phi \mid \theta)\,,
\end{equation}%
where $\lambda$ is a hyperparameter balancing the scale of flow matching loss and discriminator loss. During the training, we set $\lambda$ = 0.1.

\section{Experiments}
\subsection{Experimental Setup}
\begin{table*}[t]
\small
\renewcommand{\arraystretch}{1.2}
\centering
\begin{tabular}{c|c|c|c|c|c|c|c}
\hline
\multirow{2}{*}{Method} & \multicolumn{2}{c|}{Single Object Grasping} & \multicolumn{2}{c|}{Multi Object Grasping} & \multicolumn{2}{c|}{Burger assembly} & \multicolumn{1}{c}{Pouring}  \\ \cline{2-8}
 & I.D. & O.O.D & I.D. & O.O.D & I.D. & O.O.D & I.D. \\ \hline 
$\pi_0$ & 19/20 & 19/20 & 25/30 & 16/30 & 5/12 & 3/12&  0/20\\
\rowcolor{gray!15}
GR00T N1 & 18/20 & 13/20 & 6/30 &  8/30 & 4/12 &  3/12& 0/20\\
HAT w/ human & 17/20 & 15/20 & - & - & - & -&  2/20\\
\rowcolor{gray!15}
\acronym{} w/o human & 18/20 & 18/20 & 23/30 & 15/30 & 7/12 & 2/12& 2/20 \\
\acronym{} (Ours) & \textbf{20/20} & \textbf{19/20} & \textbf{29/30} & \textbf{30/30} & \textbf{8/12} & \textbf{7/12} & \textbf{5/20}\\  \hline
\end{tabular}
\caption{Baseline comparison results. Our method achieves the best performance among all baselines across the four manipulation tasks, under both I.D. and O.O.D. settings. We also show that training with large-scale human data improves model performance.}
\vspace{-10pt}
\label{tab:main_res}
\end{table*}

\textbf{Implementation Details.} For raw human-humanoid data, we use timestamps to synchronize episodes and process the states and actions into the human-centric representation with 240×320 images. To obtain the base \acronym{} model, we train on 8 H200 GPUs for 100k steps using 160 batch size. The weights are initialized with pre-trained checkpoint~\citep{black2024pi_0}. For post-training, we fine-tune the trained base model on a single H100 GPU for 30k steps using 10 batch size. This demonstrates one potential application of our base model to democratize egocentric manipulation training with just a single GPU.

\textbf{Robot Platforms.} For data collection and policy deployment, we use a Unitree H1 and a Unitree G1 humanoid robot. Most of the data was collected on the G1 robot. Thus, unless otherwise stated, the data and experiments are done on the G1 robot. Both robots are equipped with Inspire 5-fingered dexterous hands.

\textbf{Baselines.} We compare with 4 baseline models. $\pi_0$~\citep{black2024pi_0} is a language-conditioned flow matching model trained on many robot embodiments, which is also the initialization we use before pre-training. GR00T N1~\citep{bjorck2025gr00t} is another language-conditioned VLA using diffusion transformers. HAT~\citep{qiu2025humanoid} trains specialist policies and is thus unsuitable for pre-training or tasks that require language conditioning. Finally, \acronym{} w/o human follows the same training procedure, but without any human data in both stages.

\textbf{Experimental Protocol.} We experiment with 4 different humanoid manipulation tasks with \textit{in-distribution (I.D.)} and Out-Of-Distribution (O.O.D.) settings. The I.D. setting tests the learned skills with language, scenes, and objects that approximately resemble corresponding sequences in the robot training demonstrations. The O.O.D. setting tests configurations that are unseen in the robot training data, but may present in human data.

The tasks are illustrated in Fig.~\ref{fig:real_robot}. Objects used in these tasks are visualized in the supplementary material. Specifically,

\begin{itemize}
    \item \textbf{Single object grasping} is a sanity check task. The robot is placed in front of a table with an object and a container. The robot is tasked to pick up the object, and place it into the container. \textbf{OOD setting:} the robot is presented with objects unseen in the robot training data, different table heights, and operate in novel scenes.
    \item \textbf{Multi object grasping} is an extension of the single object grasping, where we add distractor objects. As shown in Fig.~\ref{fig:real_robot}, the robot is tasked to grasp the object. The robot must follow the language instruction and distinguish the object from distractors. \textbf{OOD setting:} the robot is presented with target objects and distractors unseen in the robot training data, different table heights, and operate in novel scenes.
    \item \textbf{Burger assembly} is intended to mimic a real-world task, where a worker at a fast food restaurant or at a food processing facility assembles a burger based on language instructions. The task is long-horizon, which involves multiple steps from using tongs to pick up ingredients specified by language, and putting the top bread. In addition, collecting on-task human data for this application can be hypothetically done by having the actual workers use wearable devices. \textbf{OOD setting:} the robot is presented with ingredients unseen in the robot training data ({\it e.g., Mozzarella cheese}) and operate in novel scenes with different table heights.
    \item \textbf{Pouring} shows the few-shot learning capability of our model. Compared to previous tasks that have hundreds of robot sequences per task, we use only \textbf{1} robot training demonstration in the bimanual pouring task, to demonstrate how \acronym{} enables few-shot robot learning.
\end{itemize}

\subsection{Evaluation}

\subsubsection{Main Experiment}

\textbf{Zero-shot language following capability from human data.} The most interesting finding is \acronym{} emerges \textit{capability to follow language instructions unseen in the robot training data}. One major weakness of existing VLAs is that they are bad at following language instructions unseen in the training data. For instance, in the multi-object grasping setting, both $\pi_0$ and GR00T N1 fail to grasp unseen objects - $\pi_0$ would randomly grasp 1 out of the 2 objects, resulting in approximately $50\%$ success rate.

On the other hand, \acronym{} is robust at following the language presented only in the human data. In the multi-object grasping experiment, the robot is capable of grasping unseen objects robustly with variations of distractors and scenes. In the burger assembly task, the robot needs to use tongs to pick up different ingredients specified by the model. Again, human data enables the model to use tools to pick up Mozzarella cheese, which is an ingredient seen only in the human data.
In sum, this capability is exciting, as it opens a door to scale up the language understanding ability of robot manipulation via egocentric human data.

\textbf{\acronym{} enables 1-shot robot data learning.} Next question we asked is - where is the boundary of such a language following capability? Can the robot learn a completely new behavior from just the human data? Empirically, the answer is \textit{no} at the current training scale. However, training with vast human data still enables few-shot learning capability. With just a single robot demonstration of the bimanual pouring task, \acronym{} achieves 5/20 success rate. We believe that the few-shot performance would further increase with larger training data scale, which may ultimately lead to zero-shot behavior.

\textbf{Human data improves overall performance on challenging task.} The burger assembly task is a challenging long-horizon task that involves tool usage, working alongside distractors, and perform multiple actions. \acronym{} outperforms baseline methods with over 100\% relative improvement on this challenging task. Notably, in the O.O.D. scenarios, we intentionally task the robot to manipulate ingredients unseen in the robot data ({\it i.e.,} red cabbage, Mozzarella cheese, and swiss cheese) to mimic special requests to food workers in the real world. In addition to language following capability discussed above, we find that the model is more robust to external disturbances such as lighting or background changes.

\subsubsection{Ablation Study}

\begin{table}[t]
\small
\renewcommand{\arraystretch}{1.2}
\centering
\begin{tabular}{c|c|c|c|c}
\hline
\multirow{2}{*}{Discriminator} & \multicolumn{4}{c}{Staged Pouring SR}  \\ \cline{2-5}
 & Right grasp & Left grasp & Pour & SR \\ \hline 
 \ding{55} &  \textbf{17/20} & 5/20 & 3/20 & 15\% \\
 \ding{51} & 16/20 & \textbf{7/20} & \textbf{5/20} & \textbf{25\%} \\ \hline
\end{tabular}
\caption{Ablation study of domain adaptation using the pouring task, which is a challenging bimanual task that can be divided into 3 stages. The success rates (SR) reported are compositional.}
\label{tab:ablation_domain}
\vspace{-1em}
\end{table}

\begin{figure}[t]
  \centering
  \includegraphics[width=\linewidth]{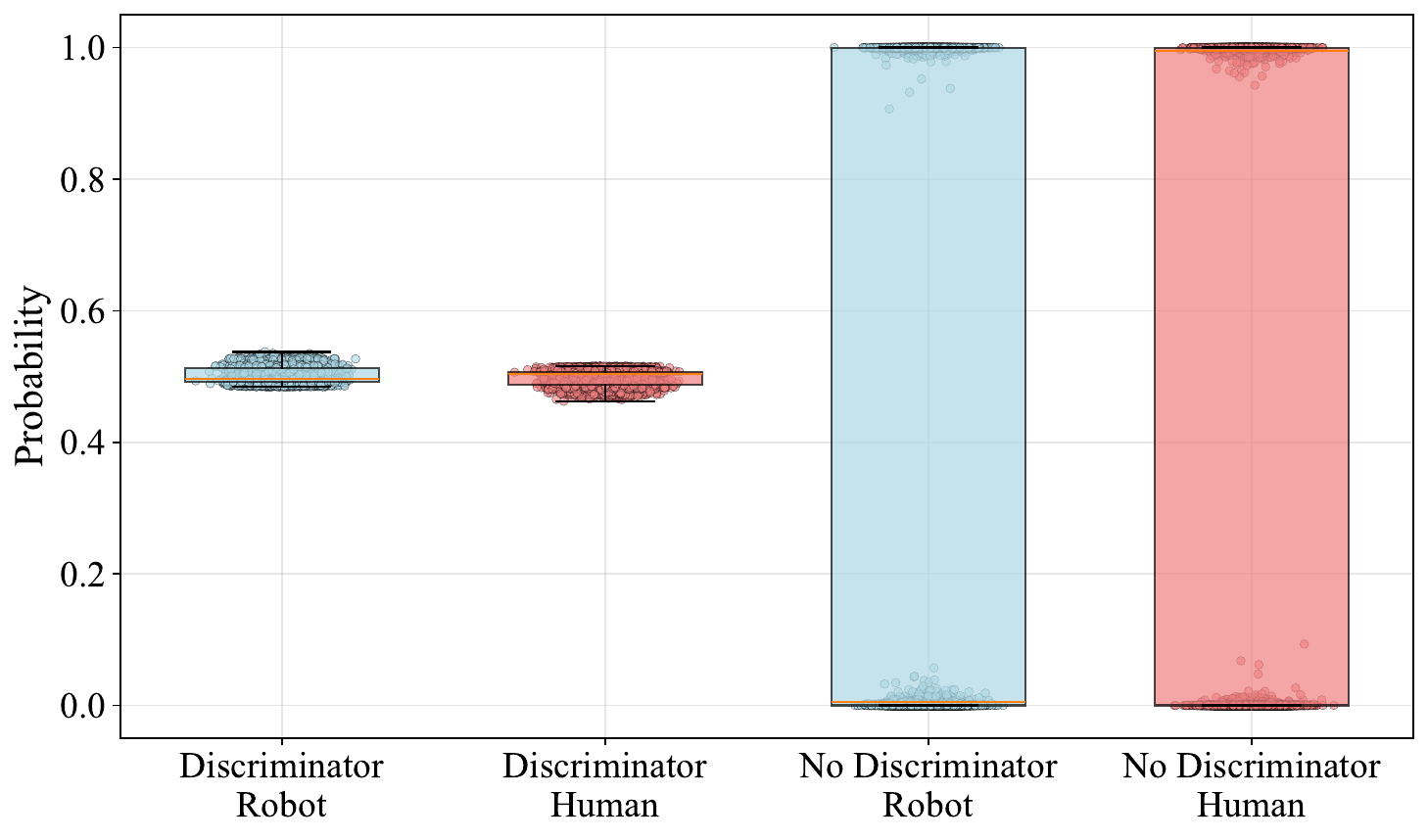}
  \caption{The probabilities produced by linear probing the intermediate features. y-axis: predicted probability of a sample being an embodiment. The model trained with a discriminator yields domain-invariant intermediate features.}
  \label{fig:boxplot}
\end{figure}

\textbf{Domain adaptation prevents model from `cheating' and helps few-shot learning.} As shown in Fig.~\ref{fig:confusion_matrix_no_dis}, the vanilla model implicitly learns to discriminate different embodiments. After adding the GRL layer, the linear probing experiment yields promising results. Fig.~\ref{fig:boxplot} suggests that linear probing can no longer discriminate embodiments, which is evident from the unconfident probabilities centered around 50\% comparable to random guesses.

As a result, domain adaptation technique improves final robot execution. Tab.~\ref{tab:ablation_domain} demonstrates improved few-shot learning capability. Without a domain discriminator, the model quickly overfits to the single humanoid demonstration. Introducing a domain discriminator encourages the shared representation to become invariant across human and humanoid domains. As a result, the humanoid can effectively leverage priors learned from human data.

\begin{figure}[t]
  \centering
  \includegraphics[width=\linewidth]{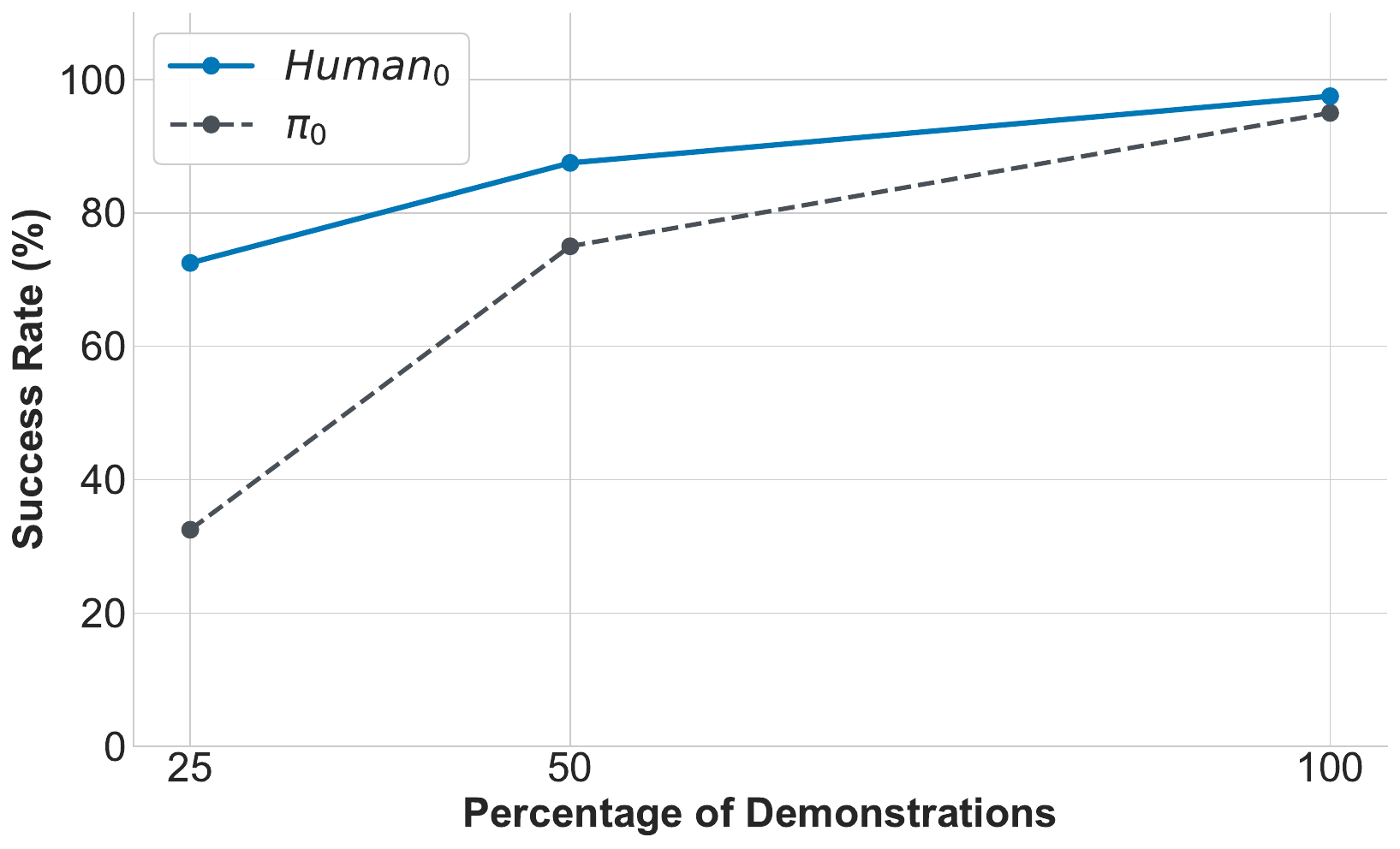}
  \caption{Performance on single object grasping. The x-axis represents the percentage of available single-object grasping robot demonstrations used in training.}
  \label{fig:data-efficiency}
\end{figure}

\textbf{Performance dynamics with robot data as a variable.} Fig.~\ref{fig:data-efficiency} shows the performance change with varying number of robot data used in training on a simple single-object grasping task. Human data can effectively regularize the learning especially in low robot data regime.

\textbf{Supplementary material.} For more results such as detailed failure analysis and videos, we encourage the readers to check out the supplementary material.

\section{Conclusion}

In this work, we presented IN-N-ON, a scalable recipe for leveraging egocentric human data through a principled taxonomy that distinguishes in-the-wild and on-task data. With \datasetname{}, a large-scale dataset comprising over 1,000 hours of diverse in-the-wild human and humanoid demonstrations and 20+ hours of task-aligned data, we enabled the training of \acronym{}. \acronym{} demonstrates several novel properties of scaling human data with language annotations. There are several interesting future directions, including further scaling of human data and testing on different robot embodiments other than humanoid robots.

{
    \small
    \bibliographystyle{ieeenat_fullname}
    \bibliography{main}
}

\clearpage
\setcounter{page}{1}
\maketitlesupplementary

\section{Appendix}
\subsection{Object visualization}
\label{app:object_viz}
\begin{figure*}[h]
    \centering
    \includegraphics[width=1.\textwidth]{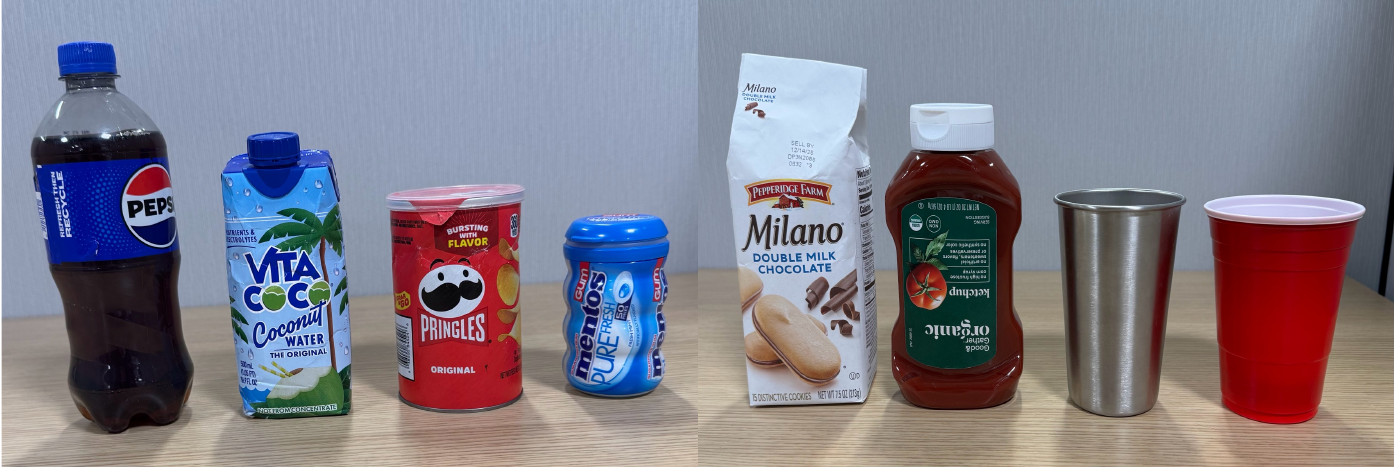}
    \caption{
        Single Object Grasping: The 4 seen objects (left) are included in the post-training, while the 4 unseen objects (right) never appear in the human demonstration data and are used to evaluate object-level generalization.
    }
    \label{fig:single_object}
\end{figure*}

\begin{figure*}[h]
    \centering
    \includegraphics[width=1.\textwidth]{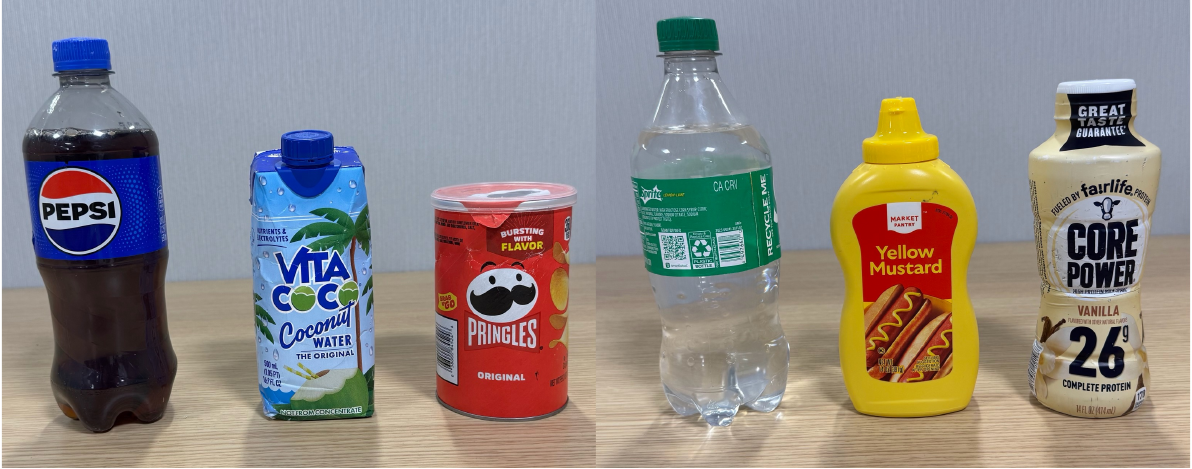}
    \caption{
        Multi Object Grasping: The 3 seen objects (left) appear in both human and robot data. The 3 unseen objects (right) are present only in the human data, together with corresponding language instructions. They are used to evaluate the zero-shot language following.
    }
    \label{fig:multi_object}
\end{figure*}

\begin{figure*}[h]
    \centering
    \includegraphics[width=1.\textwidth]{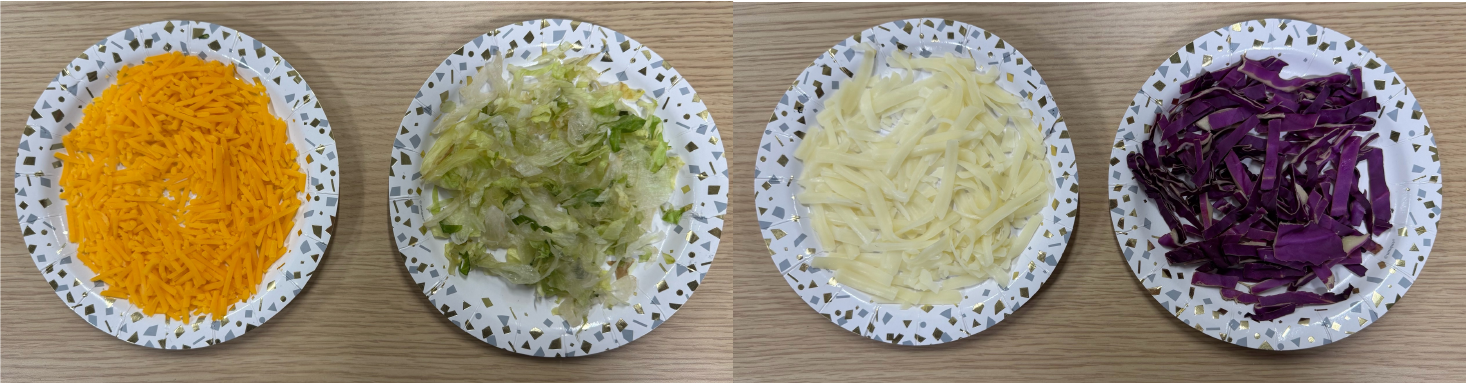}
    \caption{
        Burger Assembly: The 2 seen objects (left) appear in both human and robot data. The 2 unseen objects (right) are present only in the human data, together with corresponding language instructions. They are used to evaluate the zero-shot language following.
    }
    \label{fig:burger_assembly_object}
\end{figure*}

\begin{figure*}[h]
    \centering
    \includegraphics[width=0.6\textwidth]{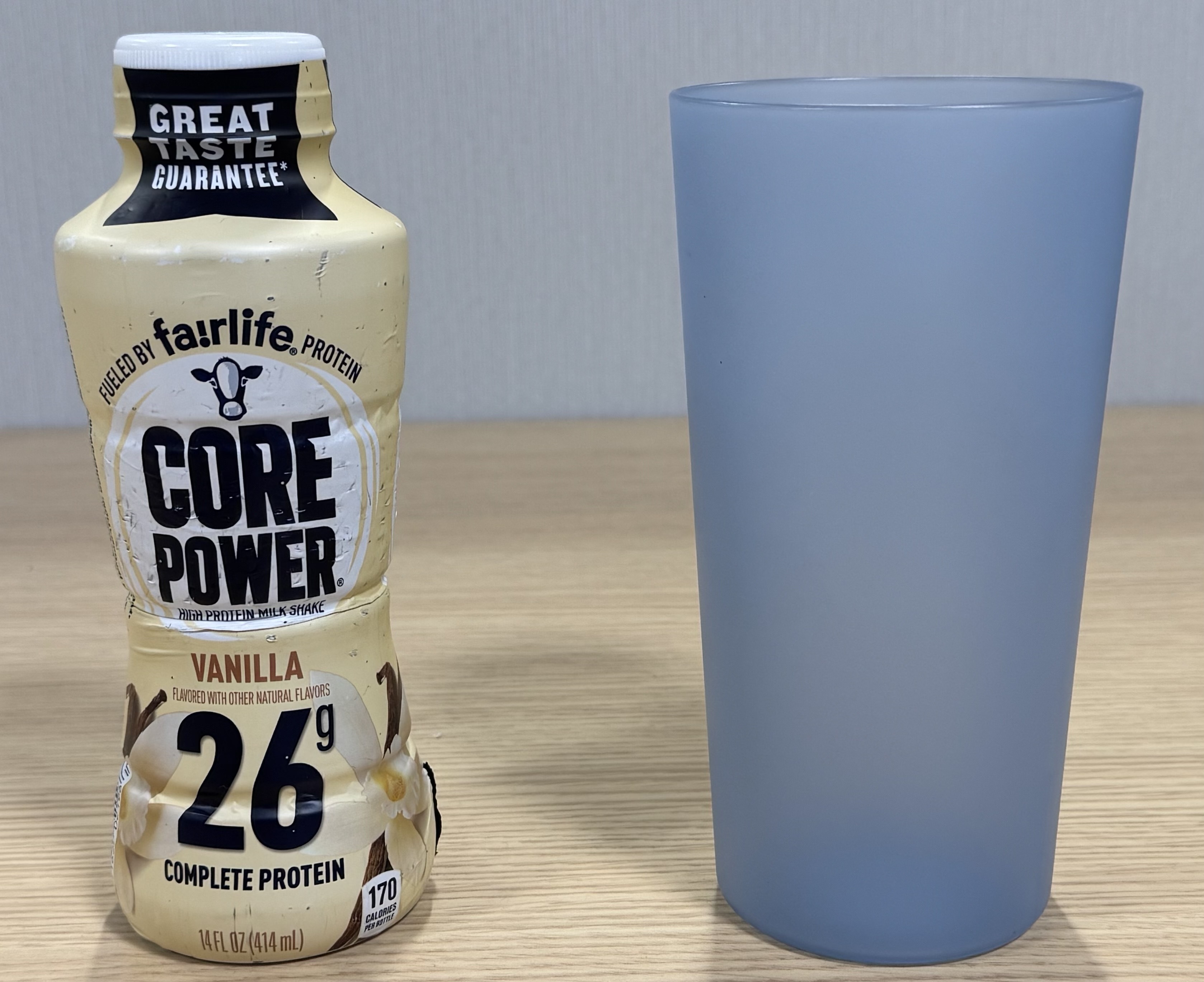}
    \caption{
        Pouring: The task uses a white protein drink bottle and a plastic cup.
    }
    \label{fig:pouring_object}
\end{figure*}

We visualized the objects used in each task in Figs~\ref{fig:single_object}, ~\ref{fig:multi_object}, ~\ref{fig:burger_assembly_object} and ~\ref{fig:pouring_object}.

\subsection{Post-training data}
We report the detailed data collected for post-training across all tasks. For each task, we count the total number of demonstrations performed by both humans and robots.
\begin{table}[h]
\centering
\label{tab:data_collection}
\begin{tabular}{l|cc}
\hline
Task & Robot & Human \\
\hline
Single Object Grasping & 120 & 2545   \\
Multi Object Grasping      & 180  & 1016  \\
Burger assembly         &  80 & 750  \\ 
\multirow{2}{*}{Pouring} 
        & 1 + 30 left grasp  & \multirow{2}{*}{727} \\
        & + 30 right grasp   &  \\
    \hline
\hline
\end{tabular}
\caption{Overview of post-training data for human and robot demonstrations. For the pouring task, we collect one initial demonstration and 30 additional demonstrations for each of the left-grasp and right-grasp configurations.}
\end{table}

\subsection{Background Ablation}
\label{sec:background_ablate}

With scaled pre-training and post-training, \acronym{} exhibits strong background generalization capabilities. To evaluate this aspect, we test the Multi-Object Grasping task under a variety of background. The results are summarized in Tab.~\ref{tab:ablation_background}.

\begin{table}[h]
\small
\renewcommand{\arraystretch}{1.2}
\centering
\begin{tabular}{l|c|c}
\hline
\multirow{2}{*}{Background} & \multicolumn{2}{c}{Multi Object Grasping}  \\ \cline{2-3}
 & I.D. & O.O.D  \\ \hline 
 White table (original) & 29/30  & 30/30 \\
 Black tablecloth & 26/30 & 29/30 \\
 Floral tablecloth & 24/30 & 28/30\\ \hline
\end{tabular}
\caption{Ablation study evaluating Multi Object Grasping performance under varying background conditions.
We report the number of successful executions over 30 trials for both I.D. and O.O.D settings. }
\label{tab:ablation_background}
\end{table}

\subsection{Multi-target language following}
We further evaluate the language-following ability of \acronym{} in multi-target settings. Specifically, we consider both in-distribution (I.D.) and out-of-distribution (O.O.D.) language instructions on the Multi-Object Grasping and Burger Assembly tasks. I.D.\ instructions are drawn from the robot data, while O.O.D. instructions involve novel language that appears only in the human data. We consider a success if it moves toward the correct object. Results are reported in Tab.~\ref{tab:language_following}. \acronym{} demonstrates robust multi-target language following on both tasks, maintaining high success rates even under O.O.D.\ instructions, indicating strong zero-shot language-following ability.

\begin{table}[h]
\small
\renewcommand{\arraystretch}{1.2}
\centering
\begin{tabular}{c|c|c|c|c}
\hline
\multirow{2}{*}{Method} & \multicolumn{2}{c|}{Multi Object Grasping} & \multicolumn{2}{c}{Burger assembly}   \\ \cline{2-5}
 & I.D. & O.O.D & I.D. & O.O.D  \\ \hline 

\acronym{} (Ours) & 29/30 & 30/30 & 10/12 & 10/12\\  \hline
\end{tabular}
\caption{Evaluation of multi-target language following for \acronym{} on the Multi-Object Grasping and Burger Assembly tasks.}
\label{tab:language_following}
\end{table}

\subsection{Failure analysis}
Despite strong performance across grasping, assembly, and pouring, Human0 still exhibits consistent failure patterns tied to perception and long-horizon control. In the different scenes or under lighting changes, the model occasionally mislocalizes objects or confuses similarly colored items, leading to false grasps or collisions. These are amplified in tasks with tool use or precise manipulation. For example, during burger assembly, small inaccuracies in early grasps often snowball into downstream misplacements that the policy cannot recover from. Similarly, in pouring tasks, failures in any of the sequential sub-stages result in the entire attempt failing.

 Additionally, while domain adaptation reduces obvious human–robot discrepancies, subtle embodiment-specific issues persist. The motion near joint limits or unstable grasps in contact-rich settings suggest that the learned representation is not fully invariant to embodiment details. These issues indicate that although large-scale human data greatly improve generalization, robust performance on precise, long-horizon manipulation still requires richer demonstrations and broader embodiment coverage.

\end{document}